%% file: main.tex
\documentclass[table]{ifacconf}
\usepackage{amsfonts}
\usepackage{amsmath}
\usepackage{graphicx}  
\usepackage{enumerate} 
\usepackage{natbib}        
\usepackage{multirow}
\usepackage{algorithm}
\usepackage{algpseudocode}
\usepackage{bm}
\usepackage{arydshln}
\input{math_commands}
\usepackage{soul}
\usepackage{xcolor}
\usepackage{colortbl} 
\definecolor{Blue}{RGB}{88, 105, 225}
\definecolor{dred}{rgb}{0.6,0,0}

\newcommand{\ka}[1]{\textcolor[HTML]{000000}{#1}}
\begin{document}
\begin{frontmatter}

\title{
RCUKF: Data-Driven Modeling Meets Bayesian Estimation
}
\thanks[footnoteinfo]{This work is supported by the University of New Mexico under the School of Engineering (SOE) faculty startup and under the Research Allocations Committee (RAC) grant \#7gr4Py.}
\author{Kumar Anurag \ \ Kasra Azizi \ \ Francesco Sorrentino \ \ Wenbin Wan} 
\address{University of New Mexico, Albuquerque, NM, USA \\
\{kmranrg, azkasra, fsorrent, wwan\}@unm.edu}

\begin{abstract}
Accurate modeling is crucial in many engineering and scientific applications, yet obtaining a reliable process model for complex systems is often challenging. To address this challenge, we propose a novel framework, reservoir computing with unscented Kalman filtering (RCUKF), which integrates data-driven modeling via reservoir computing (RC) with Bayesian estimation through the unscented Kalman filter (UKF). The RC component learns the nonlinear system dynamics directly from data, serving as a surrogate process model in the UKF’s prediction step to generate state estimates in high-dimensional or chaotic regimes where nominal mathematical models may fail. Meanwhile, the UKF’s measurement update integrates real-time sensor data to correct potential drift in the data-driven model. We demonstrate RCUKF’s effectiveness on well-known benchmark problems and a real-time vehicle trajectory estimation task in a high-fidelity simulation environment.
\end{abstract}

\begin{keyword}
Data-Driven Modeling, Reservoir Computing, Unscented Kalman Filter, Bayesian Estimation, Nonlinear Dynamics
\end{keyword}

\end{frontmatter}
%===============================================================================

\section{Introduction}
\ka{Accurate process models are essential for state estimation and control of complex vehicle systems, yet deriving them is notoriously difficult due to rapidly changing operational conditions (\cite{guo2018vehicle}).} Complex aerodynamic forces, tire-road interactions, and rapidly changing operational conditions often make purely physics-based equations incomplete or impractical (\cite{li2017dynamical}). Without a sufficiently accurate model, critical functionalities like real-time trajectory tracking, fault detection, and stability management can deteriorate. This highlights the need for more flexible and accurate modeling frameworks in vehicle dynamics and control.

\textit{Literature review.}
Bayesian estimation methods have become essential tools in engineering, especially for state estimation problems involving uncertainty (\cite{asadi2022bayesian}). For instance, the Kalman filter is designed for systems with linear dynamics and Gaussian noise (\cite{kalman1960new}). To deal with nonlinear systems, one widely used extension is the extended Kalman filter (EKF), \ka{which linearizes the dynamics at each time step}. Although effective in mildly nonlinear scenarios, the EKF’s linearization can introduce large approximation errors, degrading accuracy (\cite{gelb1974applied}). By contrast, the unscented Kalman filter (UKF) (\cite{julier1997new}) propagates a set of carefully chosen sample points through the nonlinear models, often capturing higher-order statistics more accurately. More general particle filters (\cite{arulampalam2002tutorial}) can handle highly nonlinear, non-Gaussian settings via sequential Monte Carlo sampling, but at a significantly higher computational cost. In practice, the UKF offers a strong balance between accuracy and efficiency, outperforming the EKF in many nonlinear problems while remaining much more tractable than a full particle-filter approach (\cite{wan2001unscented}). This balance, however, relies critically on having a reasonably accurate process model—an assumption that can be difficult to satisfy for \ka{strongly nonlinear} or partially known systems.

On the other hand, data‑driven methods, such as deep black‑box networks, have become popular for capturing vehicle dynamics when first–principles models are unreliable (\cite{rai2020driven,jin2021new,zhang2023review}).
% \ka{Reservoir computing in particular has recently proved effective at handling strongly nonlinear chaotic data, e.g., in the separation of superimposed Lorenz signals (\cite{krishnagopal2020separation})}. 
While these models can accommodate highly nonlinear behavior, they require large, well-curated datasets and expensive training.
Reservoir computing is proposed as an alternative capable of handling large datasets at lower computational cost \ka{(\cite{jaeger2001short,maass2004computational,pathak2018model})}.

\ka{The predictions generated by reservoir computing may degrade} sharply when the system operates outside the training envelope, which is a significant drawback for safety–critical tasks. An alternative is to embed partial physical knowledge into the learning process, e.g., by minimizing the residual of governing equations, as in physics-informed neural networks (PINNs) (\cite{cuomo2022scientific,karniadakis2021physics}). PINNs can generalize better than black‑box models when the governing equations are known, yet they still rely on gradient‑based optimization and often struggle when the physics is only partially understood or when the differential operators are stiff or chaotic.

Another research line trains a neural network, typically an LSTM or feedforward network, by backpropagation to act as the process model inside a Bayesian filter, such as the EKF, UKF, or a particle filter (\cite{tan2023vehicle,bertipaglia2024unscented}). These integrated filters can correct the network’s one‑step prediction with incoming measurements, yielding better robustness to noise than the network alone. However, \ka{this} approach inherits the heavy training cost of deep networks and still depends on careful hyperparameter tuning to avoid divergence.

Although the aforementioned works have advanced the field, none have fully solved the modeling problem for complex systems. In particular, pure data‑driven networks tend to lose accuracy when they leave the training envelope and always demand costly backpropagation; PINNs still require gradient‑based optimization and become hard to set up when the physics is only partly known; and neural \ka{networks} with feedback schemes, such as \ka{filters}, inherit the same heavy training burden, while the filter can correct the prediction only if the network is already reasonably accurate. A lightweight approach that learns quickly from data without backpropagation and still provides Bayesian error correction is therefore needed.

To address \ka{this} challenge, this paper introduces a novel framework called reservoir computing with unscented Kalman filtering (RCUKF). The key contributions of this work are summarized as follows.
\begin{enumerate} [(i)]
    \item We propose a novel architecture, RCUKF, \ka{which} leverages reservoir computing within the unscented Kalman filter.
    \item We develop a method to replace the explicit process model in the Bayesian estimation with a trained reservoir computer that captures the system dynamics from data.
    \item We introduce a two-phase workflow, comprising a learning phase for training the reservoir and an estimation phase for improving predictions with sensor measurements, thereby providing more accurate state estimates than pure data-driven methods.
\end{enumerate}

The remainder of the paper is organized as follows. Section~\ref{sec:preliminaries} introduces reservoir computing and highlights its role in learning complex system dynamics. Section~\ref{sec:rcukf} presents the RCUKF framework in detail, including the key equations that unite data-driven modeling with UKF-based estimation. Section~\ref{sec:experiments} evaluates the RCUKF framework on three nonlinear chaotic systems and on a vehicle trajectory tracking task. Finally, Section~\ref{sec:conclusion} concludes the paper.
% and outlines future research directions.

\section{Preliminaries}
\label{sec:preliminaries}

Neural networks, particularly recurrent neural networks (RNNs), are frequently employed to model time series or sequential data. However, training traditional RNNs typically requires a procedure called backpropagation through time, which can be \ka{computationally expensive} and may cause issues such as the vanishing or exploding gradient problem. While advanced models, such as long short-term memory (LSTM) or gated recurrent unit (GRU), reduce these issues, they can still require careful tuning and significant computational effort.

Reservoir computing (RC), \ka{introduced by~(\cite{jaeger2001short}) and~(\cite{maass2004computational})}, offers an alternative, more lightweight approach designed for time-series modeling and prediction. Instead of training a large network end-to-end, RC maintains a randomly connected recurrent “reservoir” of hidden neurons with fixed weights. Only the final output layer (a linear readout) is trained, simplifying the optimization to a straightforward regression step rather than full backpropagation. This design retains the reservoir’s ability to capture complex temporal patterns while drastically reducing training complexity and avoiding vanishing gradients (\cite{lukovsevivcius2009reservoir}).

\begin{figure}[ht]
    \centering
    \includegraphics[width=0.45\textwidth]{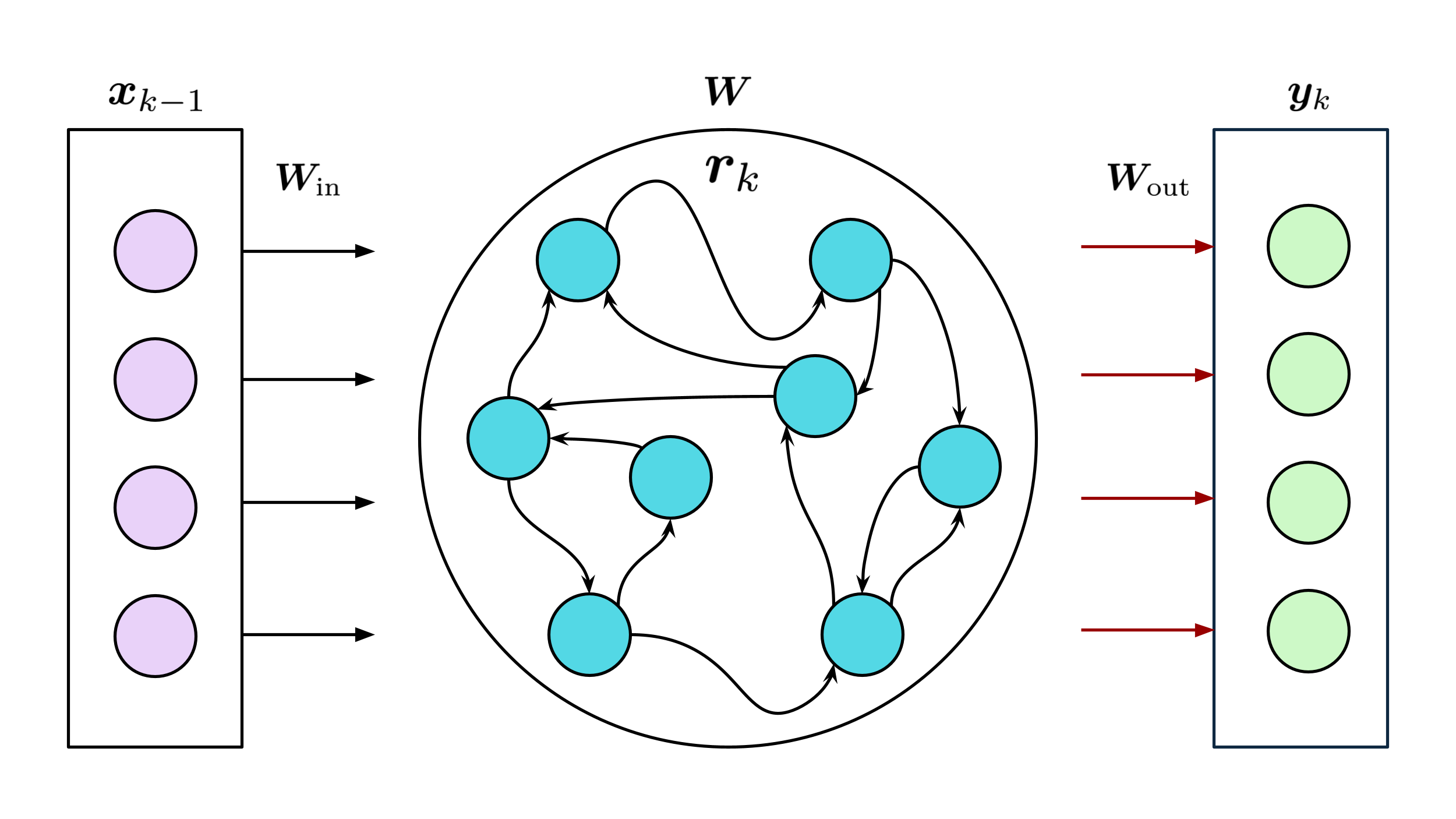}
    \caption{RC: Input \(\vx_{k-1}\) is transformed by fixed \(\mW_{\mathrm{in}}\) and \(\mW\) into reservoir, then mapped to output \(\vy_k\) via trained \(\mW_{\mathrm{out}}\).}
    \label{fig:rc}
\end{figure}

Figure~\ref{fig:rc} illustrates \ka{a standard RC framework}. The input vector $\vx_{k-1} \in \mathbb{R}^{n}$ (purple node on the left) enters the network through the input weight matrix $\mW_{\mathrm{in}} \in \mathbb{R}^{n_r \times n}$, which is randomly initialized and fixed thereafter. These inputs feed into the reservoir, comprising $n_r$ virtual neurons whose connections are defined by the recurrent weight matrix $\mW \in \mathbb{R}^{n_r \times n_r}$. The reservoir state $\vr_k \in \mathbb{R}^{n_r}$ encodes the temporal memory of the system by integrating past and current inputs. Finally, the trained output weight matrix $\mW_{\mathrm{out}} \in \mathbb{R}^{m \times n_r}$ maps the reservoir state to the output $\vy_k \in \mathbb{R}^{m}$. Because both $\mW_{\mathrm{in}}$ and $\mW$ remain fixed, the only parameters that need to be learned are in $\mW_{\mathrm{out}}$. This turns training into a straightforward linear regression problem, making RC considerably more efficient and less prone to overfitting than fully trained recurrent neural networks.
In this paper, we utilize the aforementioned beneficial properties of RC to develop a surrogate model for complex system dynamics. Specifically, the RC acts as a data-driven predictor, learning directly from historical data rather than relying on explicit equations. This learned reservoir will later replace the explicit process model typically required by traditional Bayesian estimation methods, such as the unscented Kalman filter (UKF).

The next section explains how we design the RCUKF framework by leveraging the powerful data-driven modeling capabilities of RC and the uncertainty-handling strengths of the unscented Kalman filter.

\section{RCUKF Framework}
\label{sec:rcukf}

\begin{figure}[ht]
    \centering
    \includegraphics[width=0.48\textwidth]{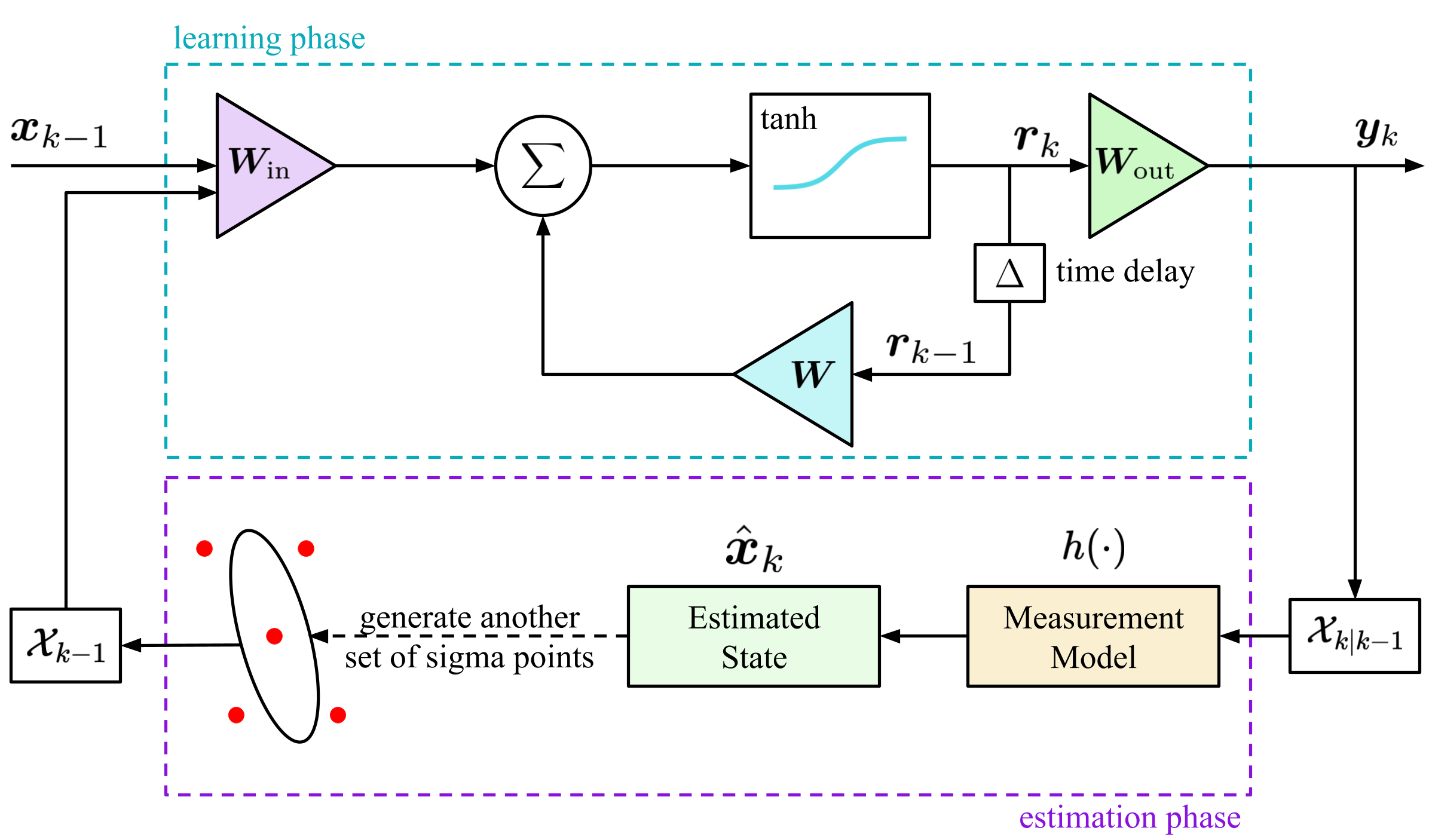}
    \caption{RCUKF: reservoir processes sigma points \(\gX_{k-1}\) with \(\mW_{\mathrm{in}}\), \(\mW\), and \(\vr_{k-1}\) to update \(\vr_k\), producing \(\gX_{k|k-1}\) via \(\mW_{\mathrm{out}}\), followed by UKF updates with measurement model \(h(\cdot)\) yielding final estimated state \(\hat{\vx}_k\).}
    \label{fig:rcukf_framework}
\end{figure}

This section presents the RCUKF framework, which integrates data-driven modeling with Bayesian filtering to enhance state estimation in complex systems, eliminating the need for an explicit analytical process model. Instead of using a predetermined function \(f(\cdot)\) to capture the dynamics of the system, RCUKF uses a reservoir computer that learns the evolution of the system directly from historical data. Real-time measurements are then assimilated through the unscented Kalman filter (UKF) to refine these predictions. The proposed RCUKF framework is executed in two phases: the \textit{learning phase} and the \textit{estimation phase}, as depicted in Figure~\ref{fig:rcukf_framework}.

During the learning phase, we collect a sequence of input states \(\bigl\{\vx_1, \vx_2, \ldots, \vx_{n}\bigr\}\). The reservoir state $\vr_k$ evolves according to the following update equation:
\begin{equation}
\label{eq:reservoir_state_update}
\vr_k = (1 - \alpha) \vr_{k-1} 
        + \alpha \tanh \bigl( \mW \,\vr_{k-1} 
        + \mW_{\mathrm{in}} \,\vx_{k-1} \bigr),
\end{equation}
where \(\alpha \in (0,1]\) is the leaking rate, controlling the \ka{effective memory window of the reservoir}, and \(\tanh(\cdot)\) is the hyperbolic tangent activation function applied element-wise. A smaller \(\alpha\) results in slower dynamics, enhancing the reservoir’s ability to retain longer-term dependencies (\cite{jaeger2007optimization}). The output is then computed as:
\begin{equation}
\label{eq:rc_predicted_state}
\vy_k = \mW_{\mathrm{out}} \,\vr_k.
\end{equation}
At the core of RC is the echo state network (ESN), which uses a large, randomly connected reservoir of neurons to transform input signals into a high-dimensional nonlinear space (\cite{maass2004computational}). The reservoir’s internal connections and input mappings are fixed, leaving only the output layer trainable. A key property of ESN is the echo state property (ESP), which ensures the reservoir states depend mainly on recent inputs rather than initial conditions. This means that the influence of the initial reservoir state fades over time. To guarantee ESP, \ka{the spectral radius of the weight matrix $\mW$, the largest absolute value of its eigenvalues, must be strictly less than one (\cite{jaeger2001short})}.

The above stability condition allows our designed reservoir to reliably represent input sequences, making it suitable for learning and predicting nonlinear system behaviors from data. The efficiency of RC arises from \ka{maintaining \(\mW_{\mathrm{in}}\) and \(\mW\) fixed,} while training only \(\mW_{\mathrm{out}}\). We define the following loss function that combines a mean squared error (MSE) term with an $\ell_2$-based regularization:
\begin{equation}
\label{eq:loss_fn}
\gL \;=\;
\underbrace{\frac{1}{N}\;\sum_{k=1}^N \bigl\|\vy_k \;-\;\vy_k^{\text{true}}\bigr\|_2^2}_{\text{Mean Squared Error}}
\;+\;
\underbrace{\delta\,\bigl\|\mW_{\mathrm{out}}\bigr\|_F^2}_{\text{Penalty Term}},
\end{equation}
where $\vy_k$ is the predicted output at time step $k$, $\vy_k^{\text{true}}$ is the target (observed) output, $N$ is the number of training samples, and $\delta>0$ is the regularization parameter. The first term is the usual mean squared error, while the second term penalizes large values in the trainable weight matrix $\mW_{\mathrm{out}}$ via its Frobenius norm, thus helping to prevent overfitting. To obtain the optimal $\mW_{\mathrm{out}}$, we solve a ridge regression problem that minimizes the \ka{loss in (\ref{eq:loss_fn})}. The closed-form solution is:
\begin{equation}
\label{eq:wout}
\mW_{\mathrm{out}}
\;=\;
\bigl(\mR\,\mR^\top \;+\;\delta\,\mI\bigr)^{-1}\,\mR\,\mY^{\text{true}},
\end{equation}
where $\mR \;=\; [\,\vr_1,\;\vr_2,\;\dots,\;\vr_n\,]$ is the matrix of collected reservoir states, $\mY^{\text{true}}=[\,\vy_1^{\text{true}},\vy_2^{\text{true}},\dots,\vy_n^{\text{true}}\,]$ is the matrix of target outputs, and $\mI$ is the identity matrix, \ka{and $\delta>0$ is the ridge-regression regularization coefficient to prevent overfitting} (\cite{lukovsevivcius2009reservoir}). Once \(\mW_{\mathrm{out}}\) is trained, the reservoir \ka{will} predict the next state solely from \(\vr_k\) using (\ref{eq:rc_predicted_state}). Because of the reservoir’s high-dimensional and nonlinear projection of input data, RC has the capability to approximate a wide range of nonlinear systems. In other words, given the training data, RC can accurately represent complex dynamical patterns. This universal approximation ability is particularly valuable when we have limited understanding or incomplete models of the system we are trying to estimate.

After learning the system’s dynamics via the reservoir, we embed it within the unscented Kalman filter to perform real-time state estimation. Consider a nonlinear system defined by:
\begin{align*}
    \vx_k &= f\bigl(\vx_{k-1}, \vu_{k-1}\bigr) + \vw_{k-1}, \\
    \vz_k &= h\bigl(\vx_k\bigr) + \vv_k,
\end{align*}
% $
%     \vx_k = f\bigl(\vx_{k-1}, \vu_{k-1}\bigr) + \vw_{k-1}, \quad 
%     \vz_k = h\bigl(\vx_k\bigr) + \vv_k,
% $
where \(\vx_k \in \R^n\) is the state vector at time \(k\), \(\vu_{k-1}\) is the control input, \(\vz_k \in \R^m\) is the measurement vector, \(f(\cdot)\) is the nonlinear process model, \(h(\cdot)\) is the nonlinear measurement model, \(\vw_{k-1} \sim \mathcal{N}(\vzero, \mQ_{k-1})\) is the process noise, and \(\vv_k \sim \mathcal{N}(\vzero, \mR_k)\) is the measurement noise. In RCUKF, we do not assume a known \(f(\cdot)\), instead, we use the trained reservoir for prediction.

Given the previous estimate \(\hat{\vx}_{k-1}\) and covariance \(\mP_{k-1}\), we generate \(2n+1\) sigma points \(\{\gX_{k-1}^{(i)}\}\) as follows:
\begin{subequations}
\begin{align}
    \gX_{k-1}^{(0)} &= \hat{\vx}_{k-1}, 
    \label{eq:sigma_point_0}\\
    \gX_{k-1}^{(i)} &= \hat{\vx}_{k-1} 
      + \Bigl(\sqrt{(n + \lambda)\,\mP_{k-1}}\Bigr)_{i}, 
      \quad i = 1, \dots, n, 
    \label{eq:sigma_point_pos}\\
    \gX_{k-1}^{(i)} &= \hat{\vx}_{k-1} 
      - \Bigl(\sqrt{(n + \lambda)\,\mP_{k-1}}\Bigr)_{\,i-n}, 
      \, i = n+1, \dots, 2n, 
    \label{eq:sigma_point_neg}
\end{align}
\end{subequations}
where {\(\lambda = \ka{\eta}^2 (n + \kappa) - n\)} is a scaling parameter, {\(\eta\)} determines the spread of the sigma points (typically \(10^{-4} \leq \eta \leq 1\)), \(\kappa\) is a secondary scaling parameter (often set to \(0\) or \(3-n\)), and \(\bigl(\sqrt{(n + \lambda)\,\mP_{k-1}}\bigr)_{i}\) is the \(i\)-th column of the matrix square root (e.g., via Cholesky decomposition). The weights for the mean and covariance are:
\begin{subequations}
\begin{align}
    W_m^{(0)} &= \frac{\lambda}{n + \lambda}, \quad 
    W_c^{(0)} = \frac{\lambda}{n + \lambda} + (1 - \eta^2 + \zeta), 
    \label{eq:weights_0}\\
    W_m^{(i)} &= W_c^{(i)} = \frac{1}{2(n + \lambda)}, 
    \quad i = 1, \dots, 2n, 
    \label{eq:weights_i}
\end{align}
\end{subequations}
where \ka{\(\zeta\)} incorporates prior distribution knowledge (often \(\zeta = 2\) for Gaussian distributions). Instead of using an explicit process model \(f(\cdot)\), each sigma point \(\gX_{k-1}^{(i)}\) passes through the trained reservoir and is updated using the following equation:
\begin{equation}
\label{eq:rcukf_propagation}
\vr_k^{(i)} = (1 - \alpha)\,\vr_{k-1}^{(i)} + \alpha\,\tanh\Bigl(\mW\,\vr_{k-1}^{(i)} + \mW_{\mathrm{in}}\,\gX_{k-1}^{(i)}\Bigr),
\end{equation}
and with the help of optimal $\mW_{\mathrm{out}}$ the propagated sigma points are then calculated as:
\begin{equation}
\label{eq:rcukf_propagation_wout}
\gX_{k|k-1}^{(i)} = \mW_{\mathrm{out}}\,\vr_k^{(i)}.
\end{equation}
The predicted state \(\hat{\vx}_{k|k-1}\) and covariance \(\mP_{k|k-1}\) are computed as:
\begin{subequations}
\begin{align}
    \hat{\vx}_{k|k-1} &= \sum_{i=0}^{2n} W_m^{(i)} \,\gX_{k|k-1}^{(i)}, 
    \label{eq:predicted_state}
\end{align}
\begin{equation}
    \begin{aligned}
        \mP_{k|k-1} &= \sum_{i=0}^{2n} W_c^{(i)} 
        \Bigl(\gX_{k|k-1}^{(i)} - \hat{\vx}_{k|k-1}\Bigr)\,\Bigl(\gX_{k|k-1}^{(i)} 
        - \hat{\vx}_{k|k-1}\Bigr)^\top \\
        &\quad + \mQ_{k-1}.
    \end{aligned}
    \label{eq:predicted_covariance}
\end{equation}
\end{subequations}
Next, the sigma points are propagated through the measurement model:
\begin{align}
    \gZ_{k|k-1}^{(i)} &= h\bigl(\gX_{k|k-1}^{(i)}\bigr), 
    \quad i = 0, \dots, 2n, 
    \label{eq:measurement_propagation}
\end{align}
and the predicted measurement, measurement covariance, and cross-covariance are:
\begin{subequations}
\begin{align}
    \hat{\vz}_{k|k-1} &= \sum_{i=0}^{2n} W_m^{(i)} \,\gZ_{k|k-1}^{(i)}, 
    \label{eq:predicted_measurement}
\end{align}
\begin{equation}
    \begin{aligned}
    \mP_{zz} &= \sum_{i=0}^{2n} W_c^{(i)} 
    \Bigl(\gZ_{k|k-1}^{(i)} - \hat{\vz}_{k|k-1}\Bigr)\,\Bigl(\gZ_{k|k-1}^{(i)} 
    - \hat{\vz}_{k|k-1}\Bigr)^\top \\
    &\quad + \mR_k,
    \end{aligned}
    \label{eq:measurement_covariance}
\end{equation}
\begin{align}
    \mP_{xz} &= \sum_{i=0}^{2n} W_c^{(i)} 
    \Bigl(\gX_{k|k-1}^{(i)} - \hat{\vx}_{k|k-1}\Bigr)\,
    \Bigl(\gZ_{k|k-1}^{(i)} - \hat{\vz}_{k|k-1}\Bigr)^\top. 
    \label{eq:cross_covariance}
\end{align}

\end{subequations}
The optimal gain can be found \ka{as} \(\mK_k = \mP_{xz}\,\mP_{zz}^{-1}\) and the state and covariance are updated as:
\begin{subequations}
\begin{align}
    \hat{\vx}_k &= \hat{\vx}_{k|k-1} 
      + \mK_k \,\bigl(\vz_k - \hat{\vz}_{k|k-1}\bigr), 
    \label{eq:state_update} \\
    \mP_k &= \mP_{k|k-1} - \mK_k\,\mP_{zz}\,\mK_k^\top. 
    \label{eq:covariance_update}
\end{align}
\end{subequations}

Using the reservoir computer for sigma point propagation, the RCUKF learns the system dynamics directly from the data, while integrating the UKF into the RC as a feedback loop to incorporate real-time sensor data and correct for any prediction drift, thereby refining the state estimates. By rescaling $\mW$ so that RC satisfies ESP, \ka{it is ensured that} the reservoir's model error is bounded. The UKF then propagates the mean and covariance without requiring Jacobians, providing a principled approximation to the Bayesian update (\cite{julier2004unscented}). This \ka{two-phase (learning \& estimation) architecture} overcomes the limitations of requiring an explicit process model and enhances state estimation.

\begin{algorithm}[t]
\caption{Reservoir Computing with Unscented Kalman Filtering (RCUKF)}
\label{alg:rcukf}
\begin{algorithmic}[1]

% --------------------------------------------------
\vspace{0.2cm}
\Statex \begin{center}\textbf{--- Learning Phase ---}\end{center}
\vspace{0.2cm}

\Statex Given a training sequence of states \(\{\,\vx_1, \vx_2, \dots, \vx_{n}\}\)
\State Randomly initialize \(\mW, \mW_{\mathrm{in}}\) with spectral radius \(\rho(\mW) < 1\) to maintain the echo state property of reservoir.

\For{\(k = 1, \ldots, n\)}
   \State Update reservoir state $\vr_k$ \Comment{Equation (\ref{eq:reservoir_state_update})}
\EndFor

\State Compute output weight $\mW_{\mathrm{out}}$ \Comment{Equation (\ref{eq:wout})}

% --------------------------------------------------
\Statex \vspace{1pt}
\Statex \begin{center}\textbf{--- Estimation Phase ---}\end{center}
\vspace{0.2cm}

\Statex Given the previous estimate \(\hat{\vx}_{k-1}\) and covariance \(\mP_{k-1}\)
\For{\(i = 0, \ldots, 2n\)}
    \State Generate sigma point \(\gX_{k-1}^{(i)}\) \Comment{Equations (\ref{eq:sigma_point_0}--\ref{eq:sigma_point_neg})}
    \State Compute weights for \(\gX_{k-1}^{(i)}\) \Comment{Equations (\ref{eq:weights_0}--\ref{eq:weights_i})}
    \State Propagate \(\gX_{k-1}^{(i)}\) using $\mW_{\mathrm{out}}$ \Comment{Equation (\ref{eq:rcukf_propagation}--\ref{eq:rcukf_propagation_wout})}
\EndFor

\State Compute \(\hat{\vx}_{k|k-1}\), \(\mP_{k|k-1}\) 
       \Comment{Equations (\ref{eq:predicted_state}--\ref{eq:predicted_covariance})}

\State Pass propagated $\gX_{k|k-1}^{(i)}$ through $h(\cdot)$ to obtain \(\hat{\vz}_{k|k-1}\), \(\mP_{zz}\), \text{and} \(\mP_{xz}\) \Comment{Equations (\ref{eq:measurement_propagation}--\ref{eq:cross_covariance})}

\State \(\displaystyle \mK_k \,\leftarrow\, \mP_{xz}\,\mP_{zz}^{-1}\)
\State Update \(\hat{\vx}_k\) and \(\mP_k\)
       \Comment{Equations (\ref{eq:state_update}--\ref{eq:covariance_update})}

\end{algorithmic}
\end{algorithm}

\section{Experiments}
\label{sec:experiments}

To assess the modeling and estimation performance of the proposed RCUKF framework, we evaluate it on three chaotic nonlinear systems, as shown in Table \ref{tab:chaotic_systems}, as well as \ka{on a} real-time trajectory-tracking task using the MIT FlightGoggles simulator (\cite{guerra2019flightgoggles}). Across all benchmarks, the RCUKF framework consistently outperforms the standard RC, where the previous prediction is fed back as input during inference, especially under noisy or limited-data conditions. \ka{All benchmarks were run on an Intel Core i9-14900KF CPU at 3.20 GHz with 64 GB RAM.}

\begin{table}[ht]
\centering
\renewcommand{\arraystretch}{1.5}
\setlength{\dashlinedash}{1pt}
\setlength{\dashlinegap}{1.5pt}
\setlength{\arrayrulewidth}{0.3pt}
\caption{Chaotic system equations and parameters.}
\begin{tabular}{|>{\centering\arraybackslash}p{0.42\columnwidth}|>{\centering\arraybackslash}p{0.42\columnwidth}|} % Same widths
\hline
\textbf{Lorenz System} & \textbf{Rössler System} \\
\hline
\begin{minipage}{\linewidth}
\vspace{3pt}
\centering
\(\displaystyle
\begin{aligned}
\dot{x} &= \sigma(y-x) + \epsilon_x,\\
\dot{y} &= x(\rho-z) - y + \epsilon_y,\\
\dot{z} &= xy - \beta z + \epsilon_z
\end{aligned}
\)
\vspace{3pt}
\end{minipage}
&
\begin{minipage}{0.88\linewidth}
\vspace{3pt}
\centering
\(\displaystyle
\begin{aligned}
\dot{x} &= -(y+z) + \epsilon_x,\\
\dot{y} &= x + ay + \epsilon_y,\\
\dot{z} &= b + z(x-c) + \epsilon_z
\end{aligned}
\)
\vspace{3pt}
\end{minipage}
\\
\cdashline{1-2}
\(
\sigma=10, \quad \rho=28, \quad \beta=\frac{8}{3}
\)
&
\(
a=0.2, \quad b=0.2, \quad c=5.7
\)
\\
\hline
\multicolumn{2}{|c|}{\textbf{Mackey–Glass Time Series}} \\
\hline
\multicolumn{2}{|c|}{
\begin{minipage}{0.88\linewidth}
\vspace{3pt}
\centering
\(\displaystyle
\frac{dx}{dt} = \beta\,\frac{x(t-\tau)}{1+x^n(t-\tau)} - \gamma\,x(t) + \epsilon
\)
\vspace{3pt}
\end{minipage}
} \\
\cdashline{1-2}
\multicolumn{2}{|c|}{
\(
\beta=0.2, \quad \gamma=0.1, \quad \tau=17, \quad n=10
\)
} \\
\hline
\end{tabular}
\label{tab:chaotic_systems}
\end{table}

In each case, we compare the RCUKF framework against a standard RC predictor \ka{(with $\rho(\mW) = 0.9$ and washout period of 100)} under two data regimes: a low-data regime ($700$ points) and a high-data regime ($10{,}000$ points). \ka{Each} dataset is divided into training and testing sets, with the first $70$\% of the points used for training and the remaining $30$\% for testing. Performance is measured using the root mean squared error (RMSE) between the true and estimated states.

\begin{figure}[ht]
    \centering
    \includegraphics[width=0.48\textwidth]{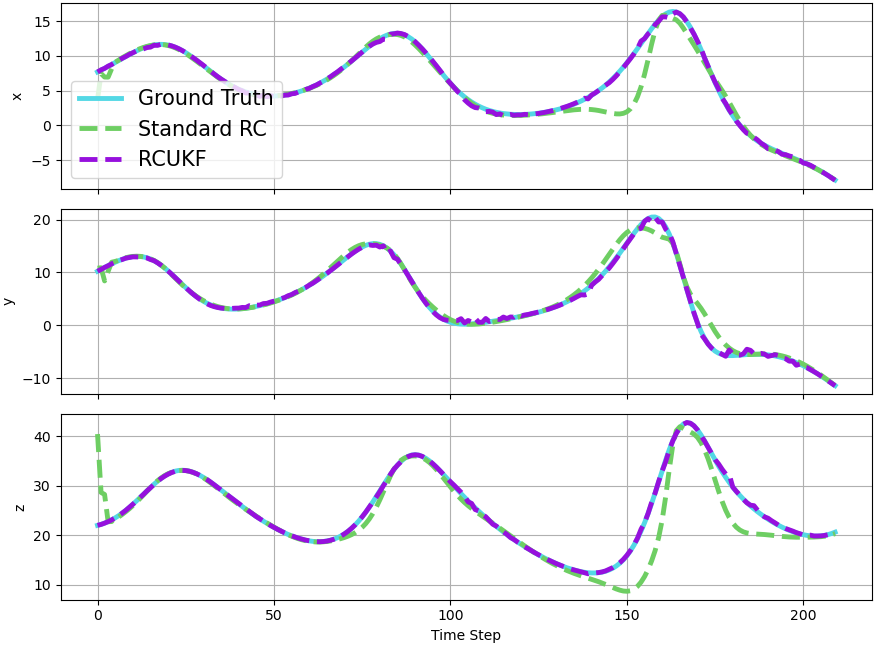}
    \caption{Standard RC vs RCUKF trajectories for next-state prediction of Lorenz System on 700 data points.}
    \label{fig:lorenz_700}
\end{figure}

\begin{figure}[ht]
    \centering
    \includegraphics[width=0.48\textwidth]{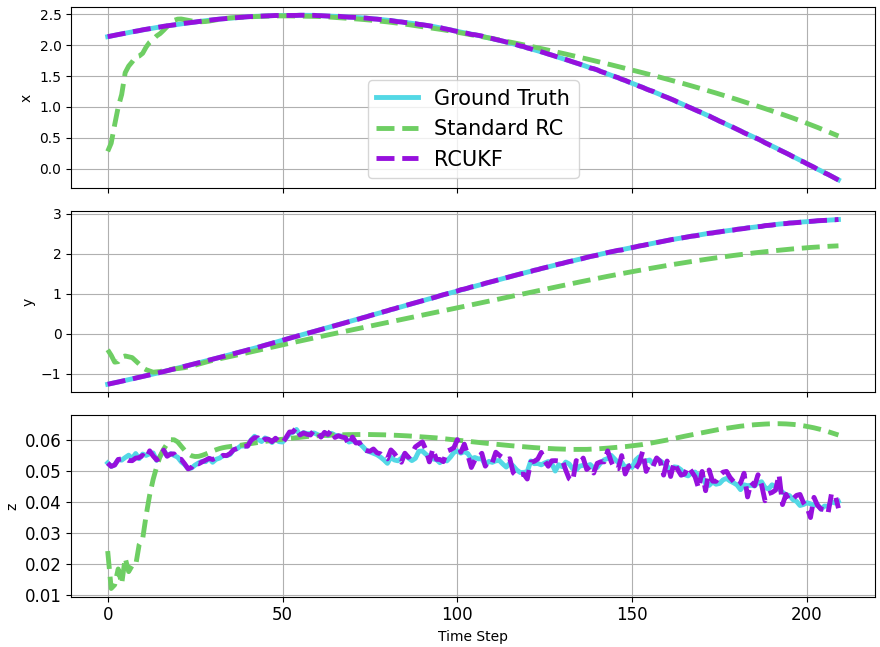}
    \caption{Standard RC vs RCUKF trajectories for next-state prediction of R\"{o}ssler System on 700 data points.}
    \label{fig:rossler_700}
\end{figure}

\begin{figure}[ht]
    \centering
    \includegraphics[width=0.48\textwidth]{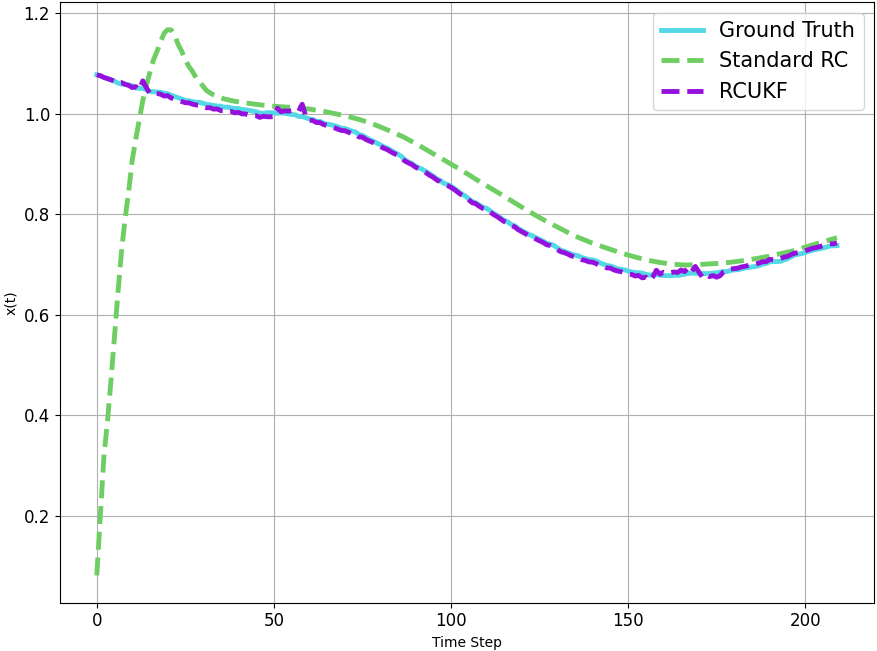}
    \caption{Standard RC vs RCUKF trajectories for next-state prediction of Mackey--Glass Time Series on 700 data points.}
    \label{fig:mackey_700}
\end{figure}

For the Lorenz system, we simulate trajectories using Euler integration with a time step of $dt=0.01$, adding independent Gaussian noise $\epsilon_x, \epsilon_y, \epsilon_z \sim \mathcal{N}(0, 0.1)$. RCUKF achieves a mean RMSE of $0.1518$ compared to $0.9318$ for standard RC, when trained on $700$ points, and maintains a strong advantage ($0.0419$ vs. $0.2125$) even with $10{,}000$ points. In the Rössler system, under the same noise conditions, RCUKF \ka{yields a lower RMSE than} standard RC ($0.1575$ vs. $0.2781$ at $700$ points; $0.0745$ vs. $0.0855$ at $10{,}000$ points), demonstrating consistent improvements even in smoother dynamics. For the Mackey–Glass series, with added noise $\epsilon \sim \mathcal{N}(0, 0.01)$, RCUKF yields a significant reduction in RMSE ($0.0032$ vs. $0.1176$ at $700$ points; $0.0014$ vs. $0.0323$ at $10{,}000$ points). Across all systems, RCUKF consistently tracks the true system states more accurately than the standard RC predictor, as visualized in Figures \ref{fig:lorenz_700}--\ref{fig:mackey_700}. A complete summary of the RMSE comparisons across all benchmarks is presented in Table \ref{tab:rmse_summary}.

\begin{table}[htbp]
\centering
\caption{RMSE Summary.}
\begin{tabular}{|l|c|c|c|}
\hline
\textbf{Metric} & \textbf{Lorenz} & \textbf{R\"ossler} & \textbf{Mackey--Glass} \\
\hline
\multicolumn{4}{|c|}{\textit{Standard RC (700 data points)}} \\
\hline
RMSE-X & 0.5829 & 0.3779 & 0.1176 \\
RMSE-Y & 0.8642 & 0.4494 &  \\
RMSE-Z & 1.3483 & 0.0070 &  \\
Mean RMSE & 0.9318 & 0.2781 & 0.1176 \\
\hline
\multicolumn{4}{|c|}{\textit{RCUKF (700 data points)}} \\
\hline
RMSE-X & 0.1628 & 0.1520 & 0.0032 \\
RMSE-Y & 0.1657 & 0.2888 &  \\
RMSE-Z & 0.1270 & 0.0316 &  \\
Mean RMSE & \textbf{0.1518} & \textbf{0.1575} & \textbf{0.0032} \\
\hline
\multicolumn{4}{|c|}{\textit{Standard RC (10,000 data points)}} \\
\hline
RMSE-X & 0.1569 & 0.1743 & 0.0323 \\
RMSE-Y & 0.2346 & 0.0448 &  \\
RMSE-Z & 0.2460 & 0.0373 &  \\
Mean RMSE & 0.2125 & 0.0855 & 0.0323 \\
\hline
\multicolumn{4}{|c|}{\textit{RCUKF (10,000 data points)}} \\
\hline
RMSE-X & 0.0370 & 0.0790 & 0.0014 \\
RMSE-Y & 0.0384 & 0.0792 &  \\
RMSE-Z & 0.0504 & 0.0653 &  \\
Mean RMSE & \textbf{0.0419} & \textbf{0.0745} & \textbf{0.0014} \\
\hline
\end{tabular}
\label{tab:rmse_summary}
\end{table}

\begin{figure}[ht]
    \centering
    \includegraphics[width=0.45\textwidth]{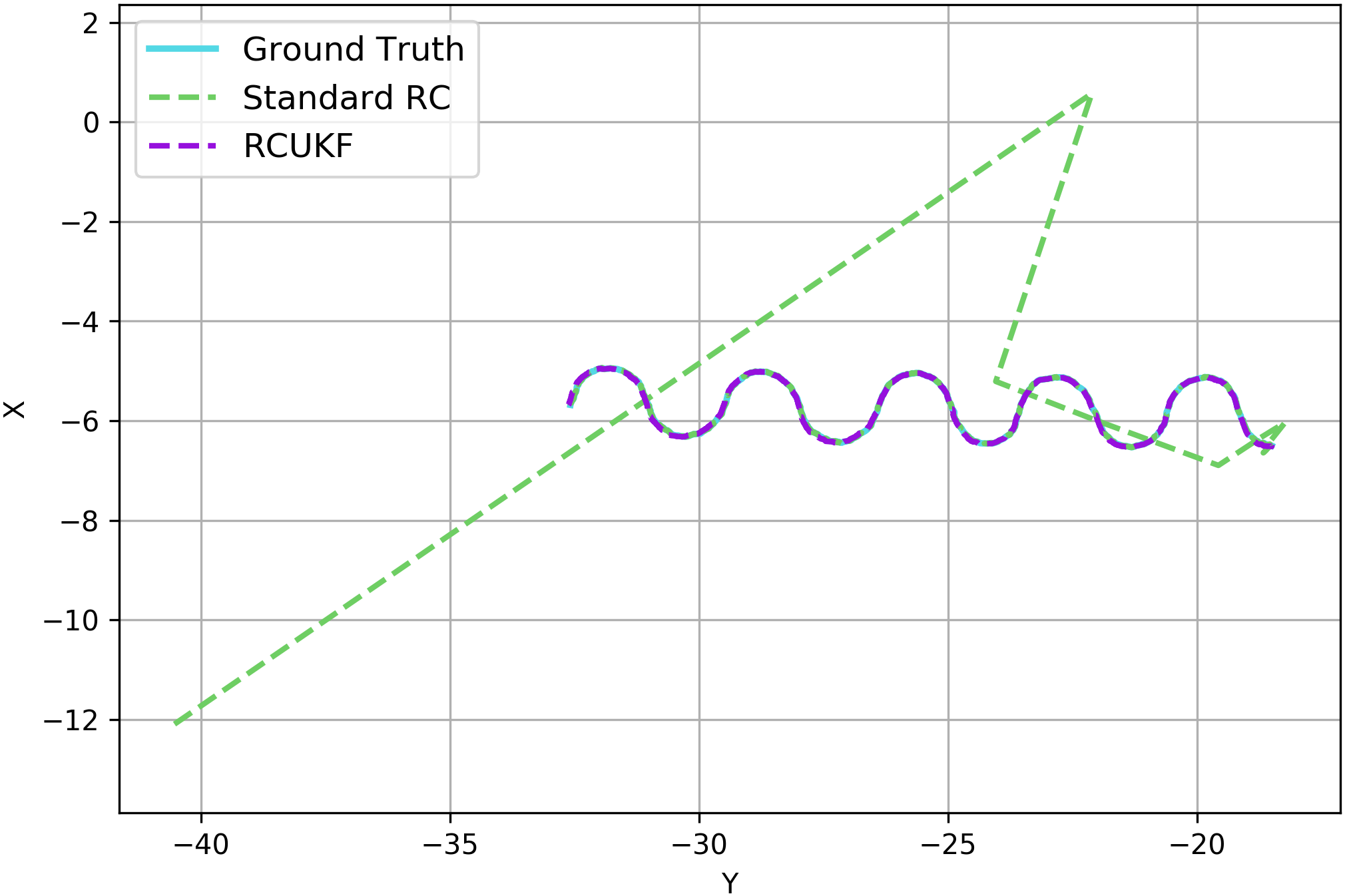}
    \caption{Real-time Lissajous tracking of a car in the MIT FlightGoggles simulation environment.}
    \label{fig:rcukf_flightgoggles}
\end{figure}

To evaluate real-time applicability in a more realistic setting, we deploy RCUKF in the MIT FlightGoggles simulator. A car is tasked to follow a double-frequency Lissajous reference trajectory:
\begin{equation*}
x(t) = R\sin(t) + \epsilon_x, \quad y(t) = R\sin(2t) + \epsilon_y,
\end{equation*}
using a PID controller for motion, where $\epsilon_x, \epsilon_y$ are process noise parameters sampled from $\mathcal{N}(0, 0.1)$ \ka{and the} amplitude $R$ is set to $1.0$. The RCUKF algorithm is applied for state estimation. To ensure \ka{realism}, we collect trajectory logs under sensor noise $\mathcal{N}(0, 0.01)$ and use $70\%$ of the data for training the reservoir component of RCUKF. The remaining $30\%$ is reserved for real-time filtering evaluation. Despite noise and complex vehicle movement, RCUKF achieves an RMSE of $0.0192$, while \ka{the standard RC achieves} an RMSE of $0.3231$. \ka{This low} RMSE score of the RCUKF framework confirms the viability for real-time applications where a fully accurate process model may not be available. Figure \ref{fig:rcukf_flightgoggles} illustrates how RCUKF closely follows the true Lissajous trajectory of the car, maintaining an accurate real-time state estimation even under noisy conditions \ka{for which the} standard RC becomes unstable.

\section{Conclusion}
\label{sec:conclusion}

This paper presented a novel framework, reservoir computing with unscented Kalman filtering (RCUKF), \ka{which} integrates data-driven modeling with Bayesian estimation for state estimation in complex dynamical systems. Instead of relying on explicit process models, RCUKF uses a reservoir computer trained on historical data to predict system evolution, while the unscented Kalman filter incorporates real-time measurements to correct predictions and manage uncertainty. We evaluated RCUKF on three chaotic benchmarks—Lorenz, Rössler, and Mackey–Glass—as well as a real-time car tracking task in the MIT FlightGoggles simulator. In all cases, RCUKF consistently achieved a lower RMSE compared to standard RC, especially under challenging conditions with limited data or noise. These results highlight the \ka{advantage} of combining data-driven dynamics learning with probabilistic state updates, supporting RCUKF's practical potential for real-time and partially known systems.

% Future work will focus on enhancing the RCUKF framework by making the reservoir's output weight $\mW_{\text{out}}$ adaptive. Instead of keeping $\mW_{\text{out}}$ fixed after training, we plan to update it online after each state estimation, allowing the reservoir to continually adapt to new system dynamics or changing environments. This adaptive learning approach is aimed to further improve the modeling accuracy and estimation resilience.

\begin{ack}
We thank Dr. Amirhossein Nazerian for \ka{helpful discussions and feedback}.
\end{ack}

\bibliography{references}   

\end{document}

%% file: math_commands.tex
\def\eqref#1{equation~\ref{#1}}
\def\1{\bm{1}}

\def\vzero{{\bm{0}}}

\def\vr{{\bm{r}}}

\def\vu{{\bm{u}}}
\def\vv{{\bm{v}}}
\def\vw{{\bm{w}}}
\def\vx{{\bm{x}}}
\def\vy{{\bm{y}}}
\def\vz{{\bm{z}}}

\def\mI{{\bm{I}}}

\def\mK{{\bm{K}}}

\def\mP{{\bm{P}}}
\def\mQ{{\bm{Q}}}
\def\mR{{\bm{R}}}

\def\mW{{\bm{W}}}

\def\mY{{\bm{Y}}}

\DeclareMathAlphabet{\mathsfit}{\encodingdefault}{\sfdefault}{m}{sl}
\SetMathAlphabet{\mathsfit}{bold}{\encodingdefault}{\sfdefault}{bx}{n}

\def\gL{{\mathcal{L}}}

\def\gX{{\mathcal{X}}}

\def\gZ{{\mathcal{Z}}}

\newcommand{\R}{\mathbb{R}}